\documentclass[conference]{IEEEtran}
\usepackage{cite}
\usepackage{graphicx}
\usepackage{xcolor}
\usepackage{booktabs}
\usepackage{url}
\usepackage{hyperref}
\usepackage{textcomp}
\usepackage{amsmath}
\usepackage{tabularx}
\usepackage{helvet}
\begin{document}

\title{When Do Tools and Planning Help Large Language Models Think? A Cost- and Latency-Aware Benchmark}

\author{\IEEEauthorblockN{Subha Ghoshal}
\IEEEauthorblockA{\textit{College of Engineering and Computer Science} \\ 
\textit{University of Michigan-Dearborn}\\
Dearborn, MI 48128, USA \\
subhag@umich.edu \\ 
} 
\and
\IEEEauthorblockN{Ali Al-Bustami}
\IEEEauthorblockA{\textit{College of Engineering and Computer Science} \\
\textit{University of Michigan-Dearborn}\\
Dearborn, MI 48128, USA \\
abustami@umich.edu}
}
\maketitle

\begin{abstract}
Modern large language models (LLMs) increasingly rely on inference-time planning and external tools to improve reasoning. We benchmark this behavior on two real-world settings: event-centric question answering over graph-structured knowledge (Event-QA) and persuasive response generation in Reddit ChangeMyView (CMV). Using LangChain and LangGraph, we compare a one-shot baseline against a plan--execute--replan agent equipped with task-specific tools (DBpedia SPARQL Protocol and RDF Query Language (SPARQL)/lookup/schema exploration, Wikipedia-focused retrieval, and topical web search). We evaluate on 60 examples each from Event-QA and CMV (3 splits of 20), and report both mean end-to-end latency and per-example token costs. We evaluate GPT-4o and GPT-4o-mini under identical workflows and report accuracy and end-to-end latency. On Event-QA, the best tool-augmented configuration improves accuracy (e.g., 47.5\% $\rightarrow$ 67.5\% for GPT‑4o) while increasing latency by orders of magnitude ($\sim$8s $\rightarrow$ $\sim$317s per example). On CMV, one-shot prompting is strongest (e.g., GPT‑4o‑mini achieves 75\% accuracy at $\sim$6s), and planning+search increases latency substantially without consistent gains. However, complex multi-tool orchestration exposes failure modes where the smaller model degrades. Overall, the findings highlight the need for task-specific, cost-aware choices of both model size and agent/tooling complexity.

\end{abstract}

\section{Introduction}
Large Language Models (LLMs) based on Transformer architectures have rapidly evolved into general-purpose systems for text understanding, generation, and reasoning \cite{vaswani2017attention,wang2024historyllm}. Beyond surface-level language fluency, recent development has emphasized the ability of LLMs \emph{to perform multi-step reasoning (planning intermediate steps and integrating external evidence)}---i.e., to perform multi-step reasoning, plan intermediate steps, and reliably integrate external information sources when answering complex real-world questions. A common and influential approach is \emph{Chain-of-Thought} (CoT) prompting, which elicits intermediate reasoning traces that can substantially improve performance on multi-step problems \cite{wei2022chain_of_thought}. Subsequent work has shown that inference-time strategies such as sampling multiple reasoning paths and aggregating consistent answers (\emph{self-consistency}) can further improve reliability without changing model parameters \cite{wang2022self_consistency}. More broadly, recent lines of research argue that \emph{test-time compute}---allocating additional deliberation or search at inference---can sometimes yield improvements comparable to or greater than scaling model size \cite{snell2024test_time_compute,muennighoff2025s1,yao2023tree_of_thoughts}.

In parallel, practical deployments increasingly rely on \emph{tool-augmented} LLMs: systems that retrieve evidence, query structured databases, or execute programs as part of producing an answer. Retrieval-Augmented Generation (RAG) combines parametric knowledge with retrieved documents to improve factuality and coverage on knowledge-intensive tasks \cite{lewis2020rag}. Agentic methods such as ReAct explicitly interleave reasoning steps with tool actions, enabling models to decompose tasks and gather missing information during inference \cite{yao2022react}. Other approaches train models to decide when and how to call external tools \cite{schick2023toolformer} or to offload computation to symbolic programs \cite{gao2022pal,chen2022program_of_thoughts}. While these methods often improve correctness and traceability, they also introduce important engineering and product trade-offs: more tool calls and longer reasoning traces typically increase latency and cost, and smaller models may struggle with the control logic needed for multi-step tool usage.
\begin{figure}
    \centering
    \includegraphics[width=1\linewidth]{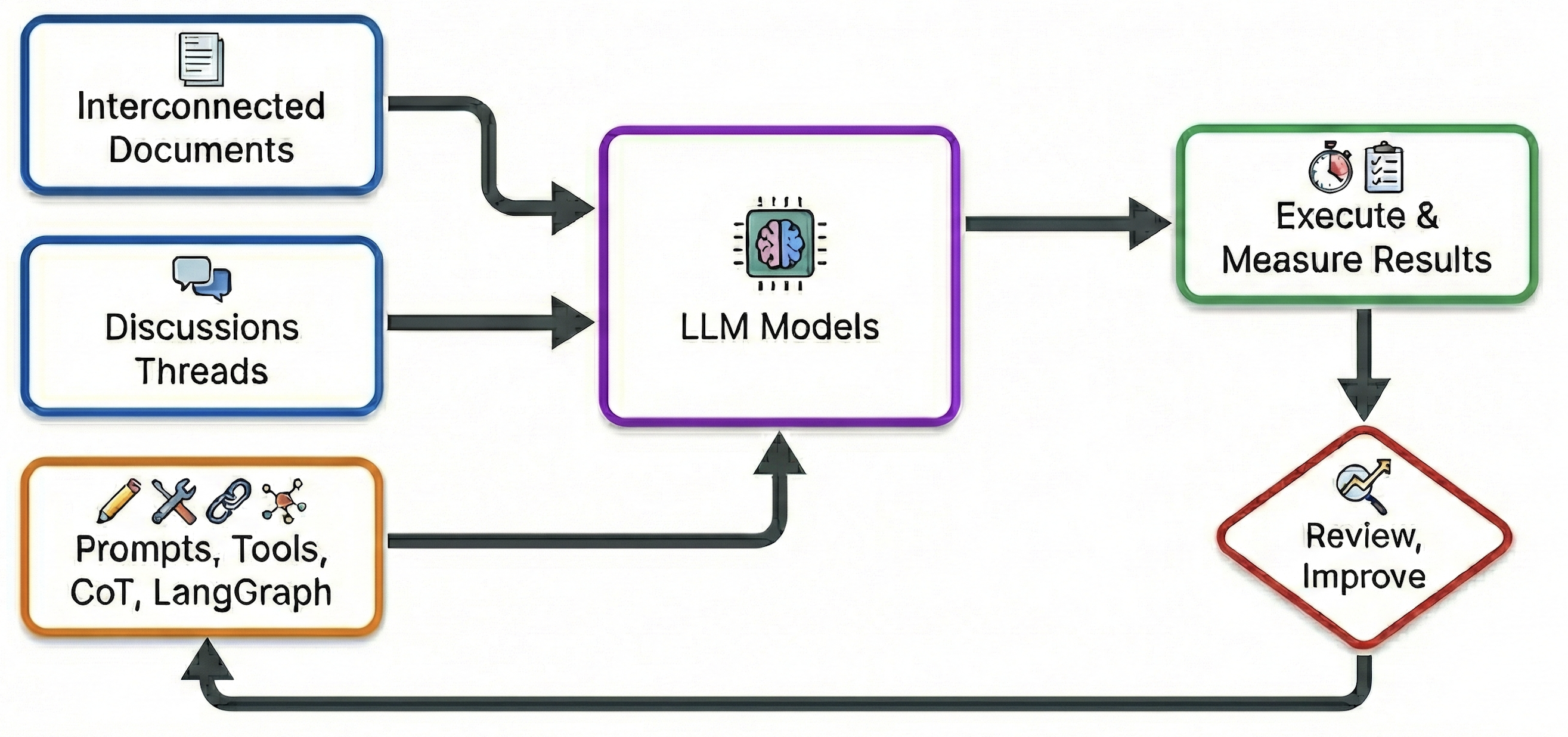}
    \caption{LLM Reasoning Evaluation Workflow, detailing the
process of planning, tool integration, and execution.}
    \label{fig:placeholder}
\end{figure}

Motivated by these trends, this paper presents an \emph{LLM Thinking Benchmark} focused on evaluating reasoning-and-tool-use effectiveness under realistic constraints. Rather than benchmarking on purely synthetic puzzles, we evaluate two real-world use cases that frequently arise in enterprise and consumer settings: (1) \textbf{event-centric question answering over graph-structured knowledge} and interconnected information sources, and (2) \textbf{argument understanding and persuasive response generation} grounded in open-domain discussions. For the first use case, we draw questions from the Event-QA dataset \cite{costa2020eventqa} and evaluate the ability of an LLM system to interpret questions that naturally map to graph queries and structured lookups. For the second use case, we use the ChangeMyView (CMV) persuasion setting derived from Reddit discussions, which has become a standard benchmark for studying persuasive interaction and argumentation dynamics \cite{tan2016winning_arguments}.

To operationalize multi-step reasoning and tool use, we implement a \emph{three-stage} CoT-style state machine (planning, execution with tool calls, and re-planning/answering) that is conceptually aligned with agentic reasoning frameworks \cite{yao2022react} and test-time deliberation methods \cite{snell2024test_time_compute,muennighoff2025s1}. For the Event-QA setting, we provide tools for querying and exploring a large public knowledge graph (DBpedia) \cite{lehmann2015dbpedia} and for searching and interpreting background text (e.g., Wikipedia). For CMV, we provide a targeted web search tool to retrieve relevant background knowledge for policy and political topics. We then compare (i) a \textbf{baseline one-shot} approach with no planning or tools against (ii) \textbf{multi-stage planning} approaches that invoke tools during inference.

A key goal of this benchmark is to inform \emph{cost-aware} system design. Modern LLM deployments face a recurring question: when does a larger, more expensive model produce sufficiently better results to justify cost and latency, and when can a smaller model (possibly with simpler tool usage) match or exceed performance? To study this, we compare a higher-capacity model (GPT-4o) against a smaller, lower-cost model (GPT-4o-mini) across both tasks and multiple prompting/tooling configurations, reporting accuracy and latency trade-offs.

We focus on three practical research questions: (RQ1) When does adding planning and tool calls improve task accuracy relative to one-shot prompting? (RQ2) What is the marginal latency and dollar cost per accuracy point gained? (RQ3) How do model size and tool orchestration complexity interact, especially in multi-tool pipelines?

In summary, this work makes three contributions:
\begin{itemize}
    \item We define a practical evaluation workflow for \emph{LLM reasoning with tools} using a plan--execute--replan structure, reflecting how many real systems are built and deployed.
    \item We benchmark one-shot prompting versus multi-stage tool-augmented reasoning on two real-world datasets (Event-QA and CMV), capturing both structured knowledge access and persuasive argumentation \cite{costa2020eventqa,tan2016winning_arguments}.
    \item We provide empirical observations on the relationship between model size, tool complexity, and cost/latency, offering guidance for cost-optimized LLM system selection.
\end{itemize}

Figure~\ref{fig:placeholder} summarizes the end-to-end evaluation workflow used throughout this benchmark.

The remainder of this paper is organized as follows: Section II reviews related work on LLM reasoning, tool use, and relevant benchmarks. Section III describes our methodology, tools, and experimental protocol. Section IV reports results, and Sections V--VI discuss findings and conclude.

\section{Related Work}
This section summarizes prior work in four areas most relevant to our benchmark: (A) reasoning and inference-time scaling, (B) tool-augmented and program-aided LLMs, (C) knowledge-graph and persuasion benchmarks, and (D) evaluation considerations for cost-aware deployment.

\subsection{Reasoning and Inference-Time Compute}
Transformer-based LLMs \cite{vaswani2017attention} have demonstrated emergent reasoning capabilities as scale increases \cite{wang2024historyllm}. Chain-of-Thought prompting \cite{wei2022chain_of_thought} provides a simple mechanism to elicit intermediate reasoning steps, often improving multi-step accuracy. However, the quality of a single generated reasoning trace can be unstable; self-consistency decoding improves robustness by sampling diverse reasoning paths and selecting the most consistent final answer \cite{wang2022self_consistency}. More generally, a growing body of work suggests that increasing inference-time compute---via additional sampling, search, or deliberation---can offer substantial gains, sometimes rivaling parameter scaling \cite{snell2024test_time_compute,muennighoff2025s1}. Tree-of-Thoughts extends CoT by exploring a search tree over partial solutions with self-evaluation, enabling backtracking and lookahead \cite{yao2023tree_of_thoughts}. These lines of work motivate our focus on comparing \emph{one-shot} answering with \emph{multi-stage} plan-and-execute pipelines that explicitly allocate additional computation at test time. Additional perspectives on inference-time scaling and recent reasoning-focused models include practitioner surveys and contemporary reasoning-model studies~\cite{raschka2025state,deepseek2025r1,li2025ttpo,wang2025underthinking,pan2025coat}.

\subsection{Tool-augmented and Program-Aided LLMs}
A complementary trend is to enhance LLMs with external tools to mitigate limitations in parametric knowledge and to improve grounding. Retrieval-Augmented Generation (RAG) retrieves supporting evidence and conditions generation on it, improving performance on knowledge-intensive NLP tasks \cite{lewis2020rag}. ReAct shows that interleaving reasoning traces with concrete actions (e.g., search) improves task completion and interpretability \cite{yao2022react}. Toolformer proposes self-supervised methods to teach language models when to call APIs and how to integrate their outputs \cite{schick2023toolformer}. Program-aided approaches offload computation to symbolic interpreters: PAL uses program synthesis to solve reasoning problems \cite{gao2022pal}, and Program-of-Thoughts prompting separates numerical computation from natural language reasoning \cite{chen2022program_of_thoughts}. Our benchmark is aligned with these approaches in that it evaluates not only final answer quality but also the practical ability of different models to \emph{use tools effectively} under multi-step control logic, where smaller models may fail due to planning or tool-invocation errors.
Several recent benchmarks evaluate LLMs as tool-using agents in interactive settings, including AgentBench and ToolBench, and realistic web environments such as WebArena. These works complement our focus by providing broad environment coverage, while our benchmark emphasizes accuracy–latency–cost trade-offs in two targeted real-world tasks.

\subsection{Benchmarks for Knowledge Graphs and Persuasion}
For structured knowledge access, question answering over knowledge graphs has long been studied, often requiring query construction and entity linking to answer compositional questions. Event-QA targets \emph{event-centric} question answering and provides query-verbalization pairs that can stress structured retrieval and reasoning over event-focused graphs \cite{costa2020eventqa}. In our work, DBpedia serves as a widely used public knowledge base and SPARQL-accessible graph for structured lookups and graph-style querying \cite{lehmann2015dbpedia}. This setup allows us to test the extent to which tool-assisted reasoning helps models translate natural language questions into structured retrieval steps.

For persuasive argumentation, the CMV community on Reddit provides a natural environment where users attempt to change an opinion holder's mind and where success is explicitly signaled. Tan et al.\ introduced and analyzed this setting, highlighting interaction and linguistic factors correlated with persuasion \cite{tan2016winning_arguments}. CMV has since been widely used for tasks such as persuasion prediction, argument quality modeling, and evidence-grounded counterargument generation. Our benchmark leverages CMV-style prompts to evaluate whether additional tool-based planning and background retrieval improve persuasive response correctness compared to direct one-shot generation.

\subsection{Cost- and Latency-Aware Evaluation}
Finally, practical system choices are increasingly shaped by the interplay between model size, inference-time computation, and tooling overhead. Recent work on test-time scaling emphasizes that allocating more compute at inference (through sampling, search, or deliberation) can be a competitive alternative to parameter scaling \cite{snell2024test_time_compute,muennighoff2025s1}. In tool-augmented systems, each additional tool call adds latency and introduces opportunities for compounding errors. Our experiments explicitly report both accuracy and latency under different planning and tool configurations, enabling cost-aware comparisons between a larger model and a smaller, cheaper alternative.

\section{Methodology}
\label{sec:methodology}

We implemented the benchmark in Python using LangChain for tool/model integration and LangGraph for deterministic orchestration of multi-step agent workflows \cite{langchain_overview,langgraph_overview,langgraph_graph_api}. We evaluated two real-world settings: (i) event-centric question answering over a knowledge graph (Event-QA) and (ii) persuasive response generation in the ChangeMyView (CMV) setting. For each dataset, we compared a one-shot baseline against a tool-augmented, multi-stage plan--execute--replan agent.

\subsection{Evaluated Approaches Per Dataset}

\subsubsection{Event-QA Approaches}
For the Event-QA dataset, we evaluated three configurations:
\begin{enumerate}
    \item \textbf{NoPlanning:} Baseline one-shot model with no tools.
    \item \textbf{Wikipedia:} 3-stage planning with  Tavily Search restricted to the Wikipedia domain.
    \item \textbf{DBpedia:} 3-stage planning with DBpedia tools (entity lookup, schema exploration, and SPARQL execution) \cite{lehmann2015dbpedia,dbpedia_sparql_endpoint,w3c_sparql11_query,dbpedia_lookup_github}.
\end{enumerate}

\subsubsection{CMV Approaches}
For the CMV dataset, we evaluated two configurations:
\begin{enumerate}
    \item \textbf{NoPlanning:} Baseline one-shot model with no tools.
    \item \textbf{PlanningSearch:} 3-stage planning with topical web search using Tavily Search. We structured retrieval across 10 topic buckets (Politics \& Civic Process, Policy Research \& Demographics, Economics \& Labor, Justice \& Law, Foreign Policy \& Security, Health \& Science, Technology \& Standards, Environment \& Energy, Immigration \& Civil Rights, Housing \& Urban Development) \cite{tavily_search_endpoint}.
\end{enumerate}

\subsection{LangGraph Approaches}
\begin{figure}[t]
    \centering
    \includegraphics[width=0.25\linewidth]{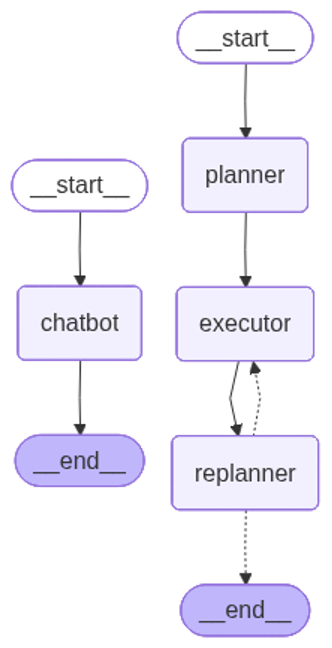}
    \caption{The two LangGraph approaches evaluated. Left: a one-shot baseline where the LLM answers directly. Right: a plan--execute--replan pipeline where the LLM plans, invokes tools during execution, and then answers or revises the plan (e.g., invokes tools for knowledge retrieval or execution).}
    \label{fig:graph}
\end{figure}

We implemented two LangGraph controller patterns, shown in Fig.~\ref{fig:graph}. The left pipeline is the one-shot baseline, and the right pipeline is the multi-stage planning agent \cite{langgraph_overview,langgraph_graph_api}.

\begin{enumerate}
    \item \textbf{Baseline one-shot (NoPlanning):} The LLM receives the question/prompt and produces an answer in a single call, without tool use.
    \item \textbf{3-stage planning (Planner--Executor--Replanner):} A three-state LangGraph agent:
    \begin{itemize}
        \item \textbf{Planner:} produces an ordered plan of steps and selects which tools to use (if any).
        \item \textbf{Executor:} completes plan steps, including structured tool invocations, and stores tool outputs as evidence.
        \item \textbf{Replanner/Answerer:} decides whether evidence is sufficient to answer; if not, revises the plan and continues.
    \end{itemize}
\end{enumerate}

\subsection{LangGraph Tools}
We exposed the following tools to the planning agent:

\begin{enumerate}
    \item \textbf{Web search tool:} returns ranked snippets from search results. For Event-QA we configured it for Wikipedia-focused retrieval; for CMV we used Tavily Search for topical background retrieval \cite{tavily_search_endpoint}.
    \item \textbf{DBpedia SPARQL query tool:} executes SPARQL 1.1 Protocol and RDF Queries against the DBpedia endpoint \cite{dbpedia_sparql_endpoint,w3c_sparql11_query}.
    \item \textbf{DBpedia resource/entity lookup tool:} resolves surface forms (e.g., entity names) into DBpedia URIs for use in SPARQL queries \cite{dbpedia_lookup_github}.
    \item \textbf{DBpedia schema explorer tool:} retrieves relevant ontology/types/properties for an entity to guide query construction (e.g., which predicates support filtering/counting) \cite{lehmann2015dbpedia}.
\end{enumerate}

\subsection{LLMs Used}
We compared two OpenAI models: GPT-4o and GPT-4o-mini \cite{openai_gpt4o_model,openai_gpt4o_mini_model}. Table~\ref{tab:model_pricing} summarizes their documented context limits and token pricing. Token prices can vary by service tier (e.g., batch vs.\ standard vs.\ priority); we cite the official pricing documentation for the values reported \cite{openai_api_pricing}. Because OpenAI does not publish official parameter counts for these models, we report commonly cited third-party estimates~\cite{claudeArtifact0ecdfb83,howarth2025_gptParameters}.

Table~\ref{tab:model_pricing} reports OpenAI’s published per-token prices for the Standard processing tier at the time of access.\cite{openai_api_pricing}.

\begin{table}[ht]
\centering
\caption{Model Specifications and Standard Pricing per 1M Token}
\label{tab:model_pricing}
\begin{tabularx}{\columnwidth}{|X|X|X|X|X|X|}
\hline
Model & Model Parameter Size & Context Token Size &
Max Output Tokens & Input Cost per million Tokens &
Output Cost per million Tokens \\
\hline
\href{https://platform.openai.com/docs/models/gpt-4o}{GPT-4o}
& Not publicly disclosed
& 128{,}000
& 16{,}384
& \$2.50
& \$10.00 \\
\hline
\href{https://platform.openai.com/docs/models/gpt-4o-mini}{GPT-4o-mini}
& Not publicly disclosed
& 128{,}000
& 16{,}384
& \$0.15
& \$0.60 \\
\hline
\end{tabularx}
\end{table}

\subsection{Experimental Protocol and Tuning}
Multi-stage tool use is substantially slower than one-shot inference, so we adopted an iterative tuning protocol to balance accuracy and runtime:

\begin{enumerate}
    \item \textbf{Prompt/controller initialization:} we created initial prompts for each configuration using small pilot subsets.
    \item \textbf{Splits:} we created three splits of 20 examples each per dataset (Event-QA stratified by question type; CMV randomly grouped).
    \item \textbf{Sequential tuning:} we tuned prompts/controller behavior on Split 1, then carried the updated configuration forward and tuned on Split 2.
    \item \textbf{Holdout evaluation:} we evaluated the tuned configuration on Split 3 without further changes.
    \item \textbf{Reporting:} final accuracy and latency were computed by averaging results from tuned Split 2 and holdout Split 3.
\end{enumerate}

\subsubsection{Event-QA Tuning}
Tuning focused on improving plan quality and reducing tool-use failure modes (e.g., entity resolution errors and incorrect SPARQL construction). We iteratively refined prompts, tool instructions, and replanning criteria to improve accuracy within the split-based workflow.

\subsubsection{CMV Tuning}
Tuning focused on improving retrieval targeting (query formulation and topical category selection) and aligning generated responses with the reference persuasive arguments. We iteratively refined prompts and retrieval settings, observing trade-offs between accuracy and inference time.

\section{Experiments}
\label{sec:experiments}

This section describes the datasets and sampling protocol, followed by quantitative results on accuracy and end-to-end inference latency for each model and approach.

\subsection{Data}
We evaluated on two labeled datasets from prior work: (i) Event-QA for event-centric question answering over knowledge graphs \cite{costa2020eventqa}, and (ii) CMV for persuasion/argumentation in good-faith online discussions \cite{tan2016winning_arguments}. Due to the high runtime cost of multi-stage tool-augmented inference (Section~\ref{sec:methodology}), we conducted experiments on fixed-size subsets.

\subsubsection{Event-QA Subset and Splits}
From Event-QA, we randomly sampled 60 question--answer pairs. We then created three \emph{stratified} splits of 20 examples each. Stratification was based on question/answer type to ensure each split contained a comparable mix of:
\begin{itemize}
    \item \textbf{Count} questions (numeric answers),
    \item \textbf{Boolean} questions (true/false answers),
    \item \textbf{Object/entity} questions (answers as entities or resources).
\end{itemize}

\subsubsection{CMV Subset and Splits}
From the CMV dataset, we randomly sampled 60 prompts and reference responses. We formed three groups of 20 examples each. Unlike Event-QA, we did not stratify CMV by a predefined question-type taxonomy due to the open-ended nature of the task.
\subsection{Evaluation Metrics}
\subsubsection{CMV Automatic Evaluation}
Because CMV responses are open-ended, we operationalize correctness using ROUGE-1 F-measure (unigram overlap) against the set of human reference responses \cite{lin2004rouge}. For each example $i$, we compute ROUGE-1 F between the generated response $\hat{y}_i$ and each reference $r \in \mathcal{R}_i$, take the best match
\begin{equation}
s_i = \max_{r \in \mathcal{R}_i} \mathrm{ROUGE}\text{-}1\_F(\hat{y}_i, r),
\end{equation}
and mark the response as correct if $s_i \ge \tau$ (fixed threshold $\tau = 0.27$). We then report accuracy as the percentage of examples marked correct:
\begin{equation}
\mathrm{Acc} = \frac{100}{N}\sum_{i=1}^N \mathbf{1}[s_i \ge \tau].
\end{equation}
Because ROUGE captures lexical overlap and may not reflect semantic equivalence or argument quality, we also recommend reporting a semantic metric (e.g., BERTScore or BLEURT) and/or a distributional metric such as MAUVE for open-ended generation.
Example LLM Instruction:
\begin{quote}
\textit{
    You are a planning assistant that updates the plan based on completed steps and their results. Use the available web search tool sparingly and avoid repeating failed searches; try alternative keywords or nearby domains from the curated list if something fails.
    If the gathered information is sufficient, provide a succinct final response that directly addresses the post and includes source domains when citing facts. If you are unsure, answer concisely with your best reasoning or say "I don't know".
    }
\end{quote}

\subsubsection{Event-QA Automatic Evaluation}
For Event-QA we have 3 different kinds of answers (Count, Boolean and Entity). We are looking for an exact match for the boolean or count values as well as for explicit dates. For names of locations or events there is some potential for variation. To assist in correctness of our comparison we perform the following:
\begin{itemize}
\item Provide explicit instructions to the LLM to help match the format provided in the reference.
\item Search for an exact substring match of the reference
\item Convert words to explicit numbers
\item Utilize a Rouge-L (Longest common substring) match of 60 percent of the reference contained in the answer to handle changes in entity name construction.
\end{itemize}
Example LLM Instruction:
\begin{quote}
\textit{
    Please answer the question as concisely as possible. Wherever possible your answer should be a single fully qualified noun, count, date, or yes or no. For nouns such as events, places, people, organizations, etc. answers with the fully qualified name and include the year of the event, etc. For count such as the number of times something occurred, answers with the number not the words. For dates, answer with the date in the format YYYY-MM-DD
    }
\end{quote}

\subsection{Results}
We report \textbf{accuracy} and \textbf{average end-to-end inference time per example}. For the multi-stage approaches, latency includes all planning, tool calls, and replanning iterations.

For the line plots (Figs.~\ref{fig:eventqa_4o}--\ref{fig:cmv_4o_mini}), the x-axis encodes both split evaluation and tuning phases:
\begin{itemize}
    \item \textbf{0.0:} Split 1 (initial configuration)
    \item \textbf{0.5:} Split 1 (after tuning on Split 1)
    \item \textbf{1.0:} Split 2 (evaluated with tuned config carried forward)
    \item \textbf{1.5:} Split 2 (after tuning on Split 2)
    \item \textbf{2.0:} Split 3 (holdout evaluation; no further tuning)
\end{itemize}
The one-shot baseline is shown as \textbf{NoPlanning}.

\subsubsection{Event-QA Results}
\begin{figure}[t]
    \centering
    \includegraphics[width=1\linewidth]{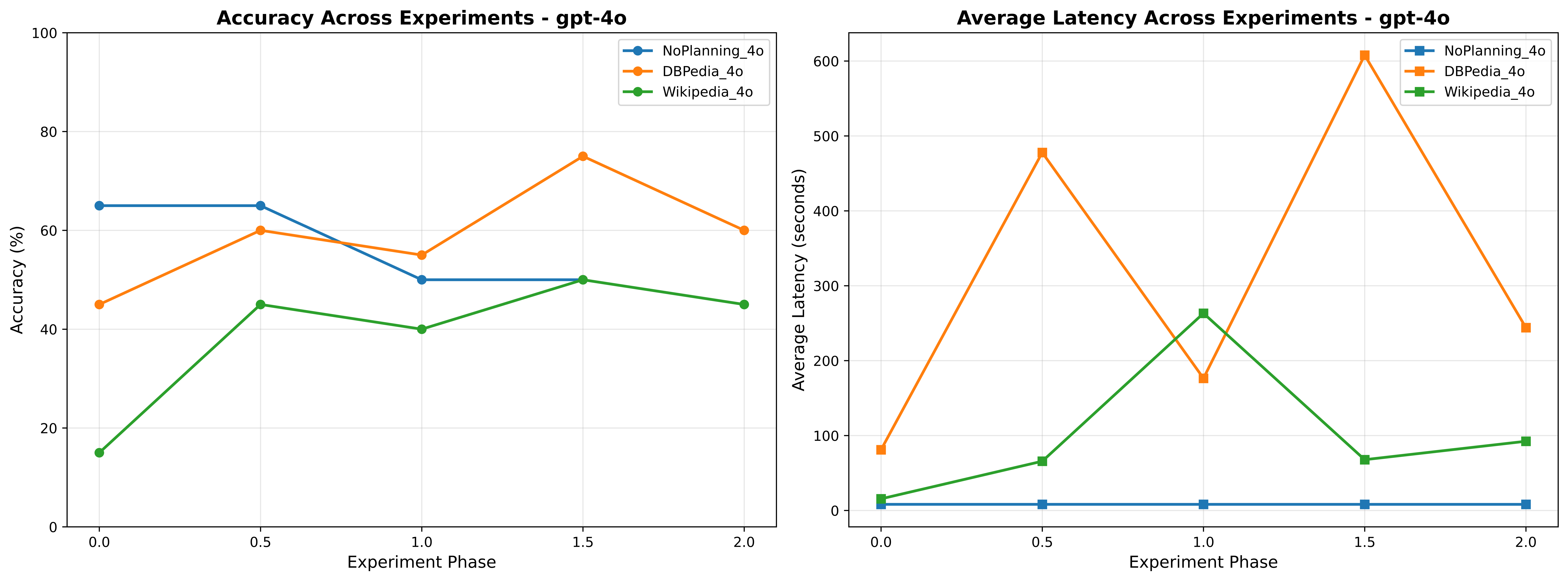}
    \caption{Event-QA - GPT-4o}
    \label{fig:eventqa_4o}
\end{figure}

\begin{figure}[t]
    \centering
    \includegraphics[width=1\linewidth]{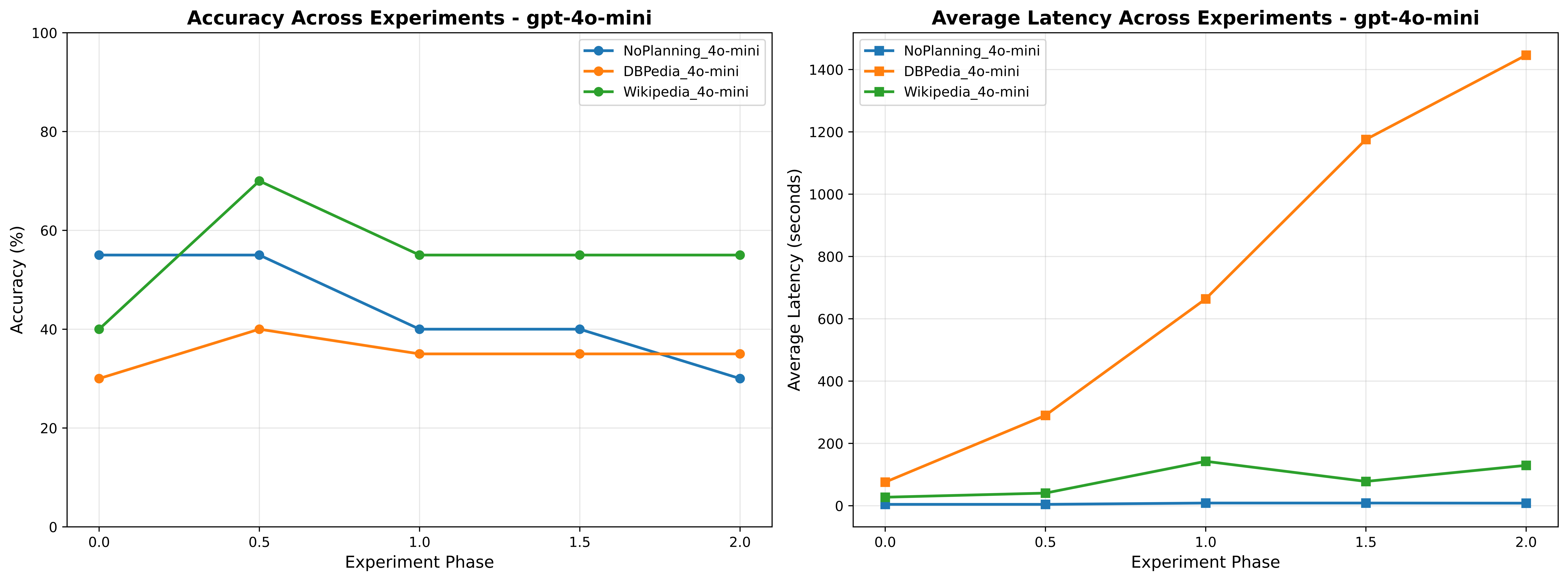}
    \caption{Event-QA - GPT-4o-mini}
    \label{fig:eventqa_4mini}
\end{figure}

Figures~\ref{fig:eventqa_4o} and \ref{fig:eventqa_4mini} report accuracy and latency across the split/tuning phases for GPT-4o and GPT-4o-mini under three approaches: NoPlanning, Wikipedia, and DBpedia. Across both models, tool-augmented 3-stage approaches generally improved accuracy over the one-shot baseline, but incurred substantially higher latency.

Overall trends observed:
\begin{itemize}
    \item \textbf{GPT-4o} performed better on \textbf{NoPlanning} and \textbf{DBpedia} than GPT-4o-mini, indicating stronger robustness to complex multi-tool, multi-step control.
    \item \textbf{GPT-4o-mini} performed best on the \textbf{Wikipedia} configuration, suggesting that simplified retrieval with lightweight reasoning can be competitive.
    \item The \textbf{DBpedia} configuration produced the highest accuracy overall (peaking at 75\% on Split 2), but also the highest average latency (hundreds of seconds per example).
\end{itemize}

\begin{figure}[t]
    \centering
    \includegraphics[width=1\linewidth]{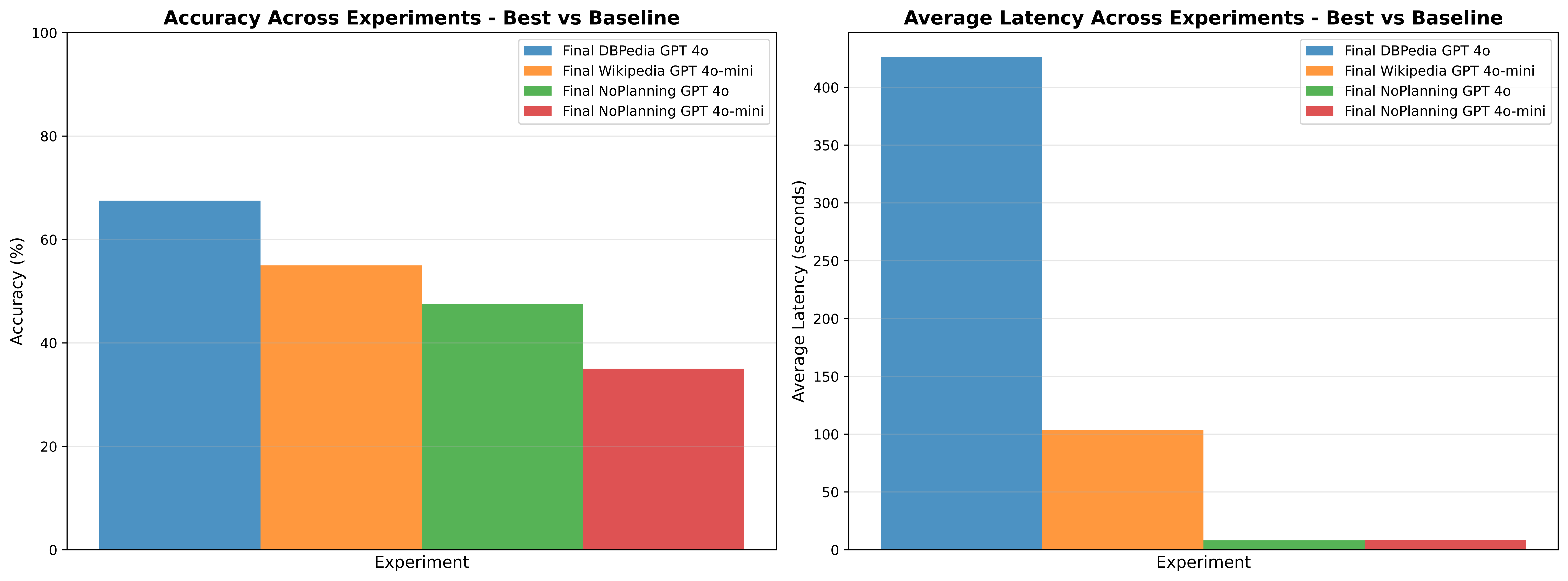}
    \caption{Event-QA - Mixed Results - Planning with DBPedia is most acurate (GPT-4o, 67.5\% Accuracy at 317s latency) while Planning with Wikipedia Search has balanced accuracy vs speed (GPT-4o-mini, 55\% Accuracy at 84s latency)}
    \label{fig:eventqa_best}
\end{figure}

Figure~\ref{fig:eventqa_best} summarizes the best overall accuracy and corresponding latency for the optimal configuration per model compared against the NoPlanning baseline. Table~\ref{tab:eventqa_model_comparison} reports the best observed split accuracy, the final reported accuracy (averaged over tuned Split 2 and holdout Split 3), and the corresponding average inference time.

\begin{table}[ht]
\centering
\caption{Event-QA: Comparison of Approach and Model Performance, Speed, and Optimal Configuration. Compares the best overall accuracy and latency for the optimal configuration per model against the NoPlanning baseline.}
\label{tab:eventqa_model_comparison}
\begin{tabularx}{\columnwidth}{|X|X|X|X|X|}
\hline
\textbf{Model} & \textbf{Best Accuracy} & \textbf{Final Accuracy} & \textbf{Avg. Inference Time} & \textbf{Optimal Configuration} \\
\hline

GPT-4o DBpedia 
& 75\% (Split 2)
& 67.5\% (Split 2 \& 3)
& \(\sim\)317 seconds
& Accurate but slow and higher cost \\
\hline

GPT-4o-mini Wikipedia
& 70\% (Split 1)
& 55\% (Split 2 \& 3)
& \(\sim\)84 seconds
& \textbf{Balanced accuracy vs speed and cost} \\
\hline

GPT-4o NoPlanning (Baseline)
& 65\% (Split 1)
& 47.5\% (Split 2 \& 3)
& \(\sim\)8 seconds
& Very fast and reasonably accurate \\
\hline

GPT-4o-mini NoPlanning (Baseline)
& 55\% (Split 1)
& 35\% (Split 2 \& 3)
& \(\sim\)7 seconds
& Lower accuracy \\
\hline

\end{tabularx}
\end{table}

\subsubsection{CMV Results}
\begin{figure}[t]
    \centering
    \includegraphics[width=1\linewidth]{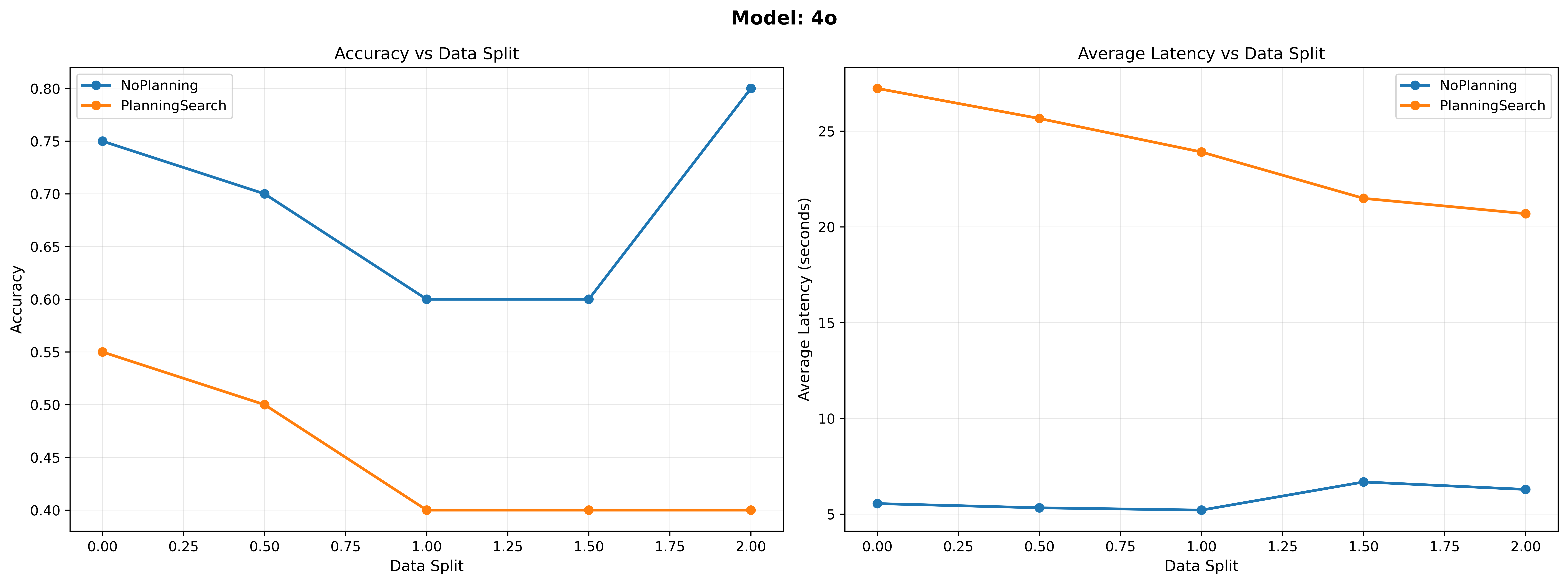}
    \caption{CMV - GPT-4o}
    \label{fig:cmv_4o}
\end{figure}

\begin{figure}[t]
    \centering
    \includegraphics[width=1\linewidth]{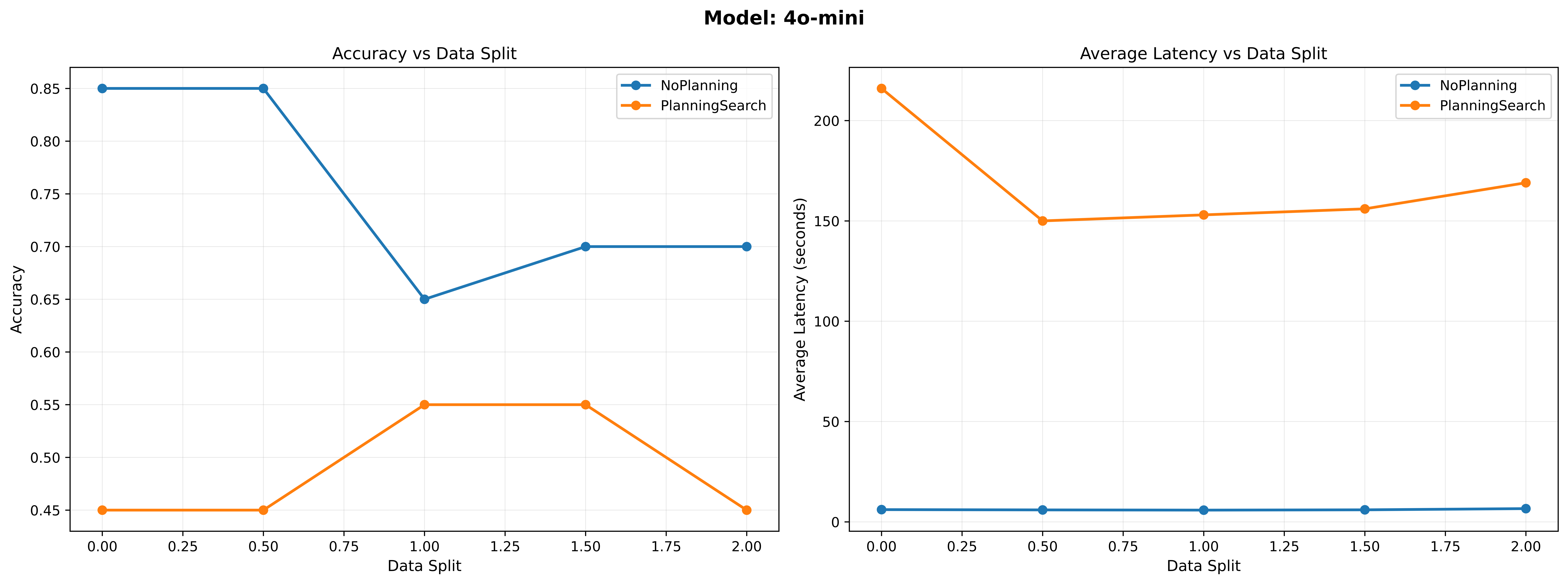}
    \caption{CMV - GPT-4o-mini}
    \label{fig:cmv_4o_mini}
\end{figure}

Figures~\ref{fig:cmv_4o} and \ref{fig:cmv_4o_mini} report accuracy and latency across split/tuning phases for CMV under two approaches: NoPlanning and PlanningSearch. In contrast to Event-QA, the one-shot NoPlanning approach achieved strong performance for both models, particularly GPT-4o-mini, while the multi-stage PlanningSearch approach often increased latency substantially without consistent accuracy gains.

Key observations:
\begin{itemize}
    \item \textbf{GPT-4o-mini NoPlanning} achieved the highest overall accuracy (up to 85\%) with consistently low latency (approximately 6 seconds).
    \item \textbf{GPT-4o PlanningSearch} maintained moderate latency (roughly 21--27 seconds) but did not outperform the one-shot baseline in final accuracy.
    \item \textbf{GPT-4o-mini PlanningSearch} exhibited the largest latency increase (approximately 150--216 seconds), making it substantially slower than both GPT-4o PlanningSearch and the NoPlanning baseline.
\end{itemize}

\begin{figure}[t]
    \centering
    \includegraphics[width=1\linewidth]{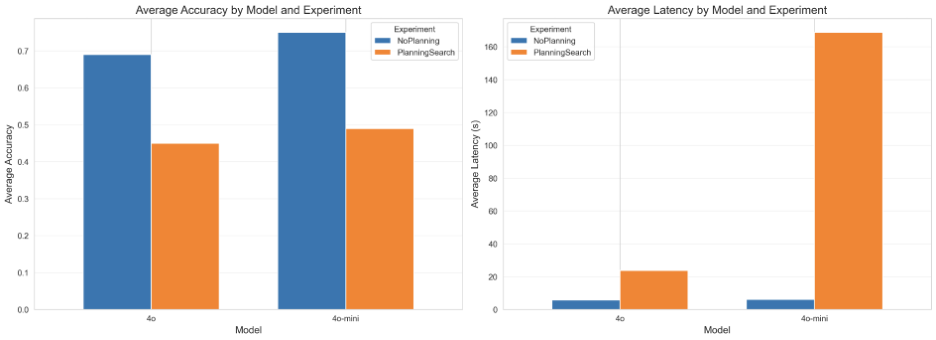}
    \caption{CMV - Best Overall - NoPlanning (GPT-4o-mini, 75\% accuracy at ~6s latency).}
    \label{fig:cmv_all}
\end{figure}

Figure~\ref{fig:cmv_all} provides an aggregate view of the best-performing configuration per model relative to the baseline, and Table~\ref{tab:cmv_model_performance_comparison} summarizes best accuracy, final accuracy, and inference-time trade-offs.

\begin{table}[ht]
\centering
\caption{CMV: Aggregate view of the best-performing configuration per model relative to the baseline, comparing accuracy, speed, and configuration.}
\label{tab:cmv_model_performance_comparison}
\begin{tabularx}{\columnwidth}{|X|X|X|X|X|}
\hline
\textbf{Model} &
\textbf{Best Accuracy} &
\textbf{Final Results} &
\textbf{Avg. Inference Time} &
\textbf{Optimal Configuration} \\
\hline

GPT-4o NoPlanning &
80\% (Split 3) &
70\% (Split 2 \& 3) &
\(\sim\)6 seconds &
Simple, fast baseline \\
\hline

GPT-4o-mini NoPlanning &
85\% (Split 1 \& 2) &
75\% (Split 2 \& 3) &
\(\sim\)6 seconds &
\textbf{Best overall trade-off} \\
\hline

GPT-4o PlanningSearch &
55\% (Split 1) &
40\% (Split 2 \& 3) &
\(\sim\)21--27 seconds &
Better than mini with planning \\
\hline

GPT-4o-mini PlanningSearch &
55\% (Split 2 \& 3) &
50\% (Split 2 \& 3) &
\(\sim\)150--216 seconds &
Computation-
ally expensive \\
\hline

\end{tabularx}
\end{table}

\section{Discussion}
In Event-QA, which involves complex data interpretation and analysis, GPT-4o appears to be better adapted at handling complex multi-stage thought processes as well as more complicated tools. In the DBpedia approach, the DBpedia queries required an element of graph schema discovery, utilizing tools provided to deduce and learn a graph ontology. GPT-4o achieved overall superior results by being able to utilize the tools and multi-stage approaches more efficiently.

GPT-4o-mini performed best on the simplified use of Wikipedia search and the ability to interpret results effectively. In contrast, GPT-4o issued more web-search calls and longer intermediate reasoning traces and interpretation of results. Given this, it appears there are cases involving simplified tool usage where GPT-4o-mini, which costs 1/16.7th as much, will provide significant value.

For the CMV benchmark, which involves argument understanding and persuasion evaluation, GPT-4o-mini demonstrated superior performance with the NoPlanning approach, achieving the highest accuracy scores (up to 85\%) while maintaining excellent inference speed. This suggests that the CMV task may not require complex multi-stage reasoning for optimal performance, and simpler one-shot prompting is sufficient when using capable smaller models.

The data reveals an interesting inverse relationship between model complexity and practical utility for this dataset:

\begin{itemize}
    \item NoPlanning approach achieved competitive or superior accuracy (60-85\%) with minimal latency (\(\sim\)6 seconds)
    \item PlanningSearch approach generally underperformed in accuracy while incurring significantly higher latency costs
    \item The 25-35x slowdown in GPT-4o-mini's PlanningSearch approach does not justify the marginal accuracy improvements
\end{itemize}

\textbf{Limitations.} Our tool-augmented settings depend on external, evolving resources (DBpedia endpoint availability and Wikipedia/web content), which can introduce nondeterminism and temporal drift. We therefore recommend caching tool outputs for released benchmark runs and reporting failure rates due to tool timeouts.

\section{Conclusion}
\label{sec:conclusion}

This paper introduced an \emph{LLM Thinking Benchmark} for studying how planning and tool use affect real-world performance under practical cost and latency constraints. We evaluated two representative settings: (i) event-centric question answering over graph-structured knowledge (Event-QA), and (ii) persuasive response generation in open-domain discussions (CMV). Across both settings, we compared a fast one-shot baseline against a plan--execute--replan agent implemented with LangGraph and task-specific tools.

Our results show that the value of "thinking" at inference time is highly task- and tool-dependent. In Event-QA, multi-stage tool augmentation improved performance relative to one-shot prompting, but introduced substantial latency. In particular, the strongest configuration (GPT-4o with DBpedia tools) achieved the best overall accuracy (Table~\ref{tab:eventqa_model_comparison}), but required orders-of-magnitude longer runtimes than the NoPlanning baseline. Meanwhile, GPT-4o-mini paired with Wikipedia-style retrieval provided a competitive accuracy--latency trade-off, suggesting that for some structured QA workloads, simpler retrieval and lighter-weight reasoning can be cost-effective.

In contrast, for CMV the simplest approach was consistently strongest: the NoPlanning baseline (especially with GPT-4o-mini) achieved the best overall accuracy with minimal latency (Table~\ref{tab:cmv_model_performance_comparison}). The PlanningSearch configuration often increased runtime substantially without improving final accuracy, indicating that additional tool calls and deliberation can be counterproductive on tasks where a model's internal priors already align well with the expected response style and where retrieval adds noise or distracts from argument quality.

Taken together, these findings suggest a practical deployment heuristic: start with a low-latency one-shot baseline (often with a smaller, cheaper model), add retrieval and planning only when the task requires structured evidence access or multi-hop composition, and escalate to a larger model primarily when tool orchestration and complex multi-step control become failure points. Future work should evaluate larger samples, additional model families, stronger automatic evaluation for open-ended persuasion, and more detailed error taxonomies for tool-use failures (e.g., entity linking vs.\ schema discovery vs.\ query formulation).

\bibliographystyle{IEEEtran}
\bibliography{references}

\end{document}